\documentclass{article}
\usepackage{ijcai18}

\usepackage{times}
\usepackage{soul}
\usepackage{url}
\usepackage[hidelinks]{hyperref}
\usepackage[utf8]{inputenc}
\usepackage[small]{caption}
\usepackage [english]{babel}
\usepackage [autostyle, english = american]{csquotes}

\usepackage{epigraph}

\usepackage{tabularx,ragged2e}
\usepackage{booktabs}
\usepackage{caption}
\usepackage{subcaption}
\usepackage{adjustbox}
\usepackage{graphicx}
\usepackage[utf8]{inputenc}
\usepackage{csquotes}
\usepackage{amssymb}

\usepackage{tabularx,ragged2e}
\usepackage{booktabs}

\usepackage{adjustbox}

\usepackage{xcolor}

\title{Machine Learning Interpretability: A Science rather than a tool}

\author{
Abdul Karim$^1$,
Avinash Mishra$^1$,
MA Hakim Newton$^1$
Abdul Sattar$^1$
\\ 
$^1$ Institute of Integrated and Intelligent Systems, Griffith University, Queensland, Nathan, Australia\\
abdul.karim@griffithuni.edu.au
}

\begin{document}

\maketitle

\begin{abstract}
The term ``interpretability" is oftenly used by machine learning researchers each with their own intuitive understanding of it. There is no universal well agreed upon definition of interpretability in machine learning. As any type of science discipline is mainly driven
by the set of formulated questions rather than by
different tools in that discipline, e.g. astrophysics is
the discipline that learns the composition of stars,
not as the discipline that use the spectroscopes.
Similarly, we propose that machine learning interpretability should
be a discipline that answers specific questions
related to interpretability. These questions can be of
statistical (associational), causal and counterfactual nature. Therefore,
there is a need to look into the interpretability
problem of machine learning in the context of
questions that need to be addressed rather than
different tools. We discuss about a hypothetical interpretability framework driven by a question based scientific approach rather than some specific machine learning model. Using a question based notion of interpretability, we can step towards understanding the science of machine learning rather than its engineering. This notion will also help us understanding any specific problem more in depth rather than relying solely on machine learning methods.

\end{abstract}

\section{Introduction}

A dramatic hype in the success of machine learning (ML) models achieving the human level performance in many different areas led to a great need of deploying these systems in real world applications. The ongoing research aims to produce autonomous end to end systems which will perceive, observe the surrounding, learn from the data and surrounding, and take a decision on training data observation. However, the employability of these systems is greatly limited by the model’s inability to explain the decisions. In fact, the article 13 of General Data Protection Regulation (GDPR) mentions about the right to be informed of each end user of the data. This implies that the models which use personal data of the users to make some decisions about the relevant person, for instance in getting a bank loan or deciding if the person to be given a specific treatment or not, the user has a full right of understanding how and why the model came across a specific decision. So the issue of interpretability is very important to be considered in future machine learning algorithms. 
``Single evaluation metric" is an incomplete description of the most real-world tasks which is the basis for all machine learning techniques. For instance, our goal in lending a loan money is to reduce the loan default ratio as well as not to discriminate against anyone on the basis of the race or area they are living in. Mostly machine learning techniques only optimize the loan default metric and does not care about the other factors like discrimination. Therefore, there is a need to include other important factors in optimization of ML techniques one of which is interpretability. ML techniques are frequently used in scientific fields to classify or predict, so they must answer the ``how" and ``why" questions to be coherent with the science goals. Interpretability will greatly help us in reducing data bias, making safe ML algorithms, debugging the safety holes and managing social interactions. There is an overview of making the traditional classification more comprehensible and discussing the fact that interpretability cannot be defined monotonically but rather it is a multi-dimensional concept\cite{l22}
\subsection{Interpretability if seen from the ``model" point of view}
Survey of the literature for machine learning model interpretability reveals that there is no agreed upon universal definition for it, and people use it in their own context and for the specific type of application or model. The interpretability refers to multiple concepts and contexts. In terms of machine learning, interpretability can be divided into two streams, either the model is interpretable itself or some other technique is applied on the black box model to extract interpretations. Interpretable models are those which are intrinsically interpretable in terms of the weights or parameters. We often find a claim in the literature survey for machine learning that linear models are more interpretable than the deep learning models\cite{l2}. This claim is very specific to the data with small number of features. In case of high dimensional data sets, which is usually the real world scenario, even the linear model will be uninterpretable or the accuracy level will be low as compared to the deep learning models. Moreover, the coefficients of the linear models only have an interpretation under very strict assumptions like linear relationship, multivariate normality, no or little multicollinearity, no auto-correlation and homoscedasticity. If in any model, any of these assumptions are not satisfied, that we cannot conclude any association based interpretation of the model parameters. Logistic regression is the non-linear version of linear regression used for classification problems. The interpretation of logistic regression always comes with the clause that all other features stay the same. Though the model itself is considered interpretable, yet it suffers from the poor classification performance. Decision tree is also considered as one of the most important interpretable model which is very intuitive and covers interactions between the features. As it is based on hard splits, so it is very inefficient to handle linear relationships between the feature and the target variable. This leads to lack of smoothness and instability. Moreover, to achieve high performance, we need to use many trees instead of one tree, which eventually makes it hard to interpret it again.

\subsection{Interpretability if seen from the ``post-hoc" point of view}
Post-hoc interpretations are also called model-agnostic methods in which the interpretations are generated from the black-box model after training. Usually, data is generated from some real world problem which is then used to create fitting function via any type of black-box method. One of the post hoc interpretability approach is training one model for predictive performance and another model to generate the interpretations of the decisions made by the model\cite{l12}. There is another approach of generating post-hoc interpretation by visualizing the learnt states or parameters after training \cite{l14},\cite{l15},\cite{l19},\cite{l20},\cite{l21}. It can be very difficult to describe the overall mapping done by the neural network during training, so some papers attempt to explain the local information in the neural network \cite{l16},\cite{l17}. It should be noted that these types of interpretation can be misleading as they are constrained to the local regions only. There is one other attempt called ``Local Interpretable Model-agnostic Explanations (LIME)" to explain the decisions locally near a particular region by learning a separate sparse linear model\cite{l5}. LIME is mostly applied to text and image datasets. Moreover, there is too much of wiggle room for optimization in order to obtain interpretations for a specific task.

\subsection{Interpretability if seen from the ``causality" point of view}
Few researchers tailor the causal relations of input features. They emphasize on the distinction of causal and prediction components of their models\cite{l6}. In real world applications like medical or some other area, our real goal is to obtain potential causal associations but the optimization goal of most of the supervised machine learning is minimizing the mean square error which can be easily achieved using only correlative associations \cite{l7}. The associations learned by the supervised machine learning algorithms are not always guaranteed to reflect the causal relationships. Therefore, there could be a casual hypothesis generated from the models which can be tested experimentally for verification\cite{l8}. Regression trees and Bayesian neural networks are considered a good candidates for this type of hypothesis generation. Pearl presents a very detailed overview of inferring causal relationships from the data\cite{l9}. These methods mainly rely on strong assumption of some prior knowledge about the data.

\section{Why current ML techniques are non-interpretable}

If we examine the current ML techniques, we find the following four causes which makes them black-box in nature at different levels.
\begin{description}

  \item[$\bullet$ Incompleteness in the problem formalization] Incompleteness in the problem formalization in ML is one of the main cause of their limited interpretability\cite{doshi2017towards}. Usually, a single metric like a classification accuracy is used to for optimization which does not completely describe the real world problem. For certain problem, It is not enough to get the answer of `` the what". We need an explanation that how the model came across certain decision ``the why" because a correct prediction is only a partial solution to the problem. Interpretability is one of the dimensions of formalizing a problem which is very hard to quantify. That is why, it is usually not included in the optimization with the accuracy. 
  \item[$\bullet$ Model opaqueness] In order to describe the opaqueness of the model, we will take a deep neural network (DNN) as an example. After training a DNN, a very big weight or parameter matrix is obtained . This weight matrix is nothing but a big pile of incomprehensible noise which makes the DNN opaque. They are excellent in mapping a function from to output, but worse in understanding the context of the data they are handling. That is why, the weight matrix after training has no intuitive meaning and it is very hard to trace back the parameters to identify the important features.
 
  \item[$\bullet$ Local and global comprehension]Local comprehension deals with the local regions of the conditional distributions. The ML techniques are lacking in local comprehension in a way that they can not explain the reason of a particular prediction. Similarly, global comprehension which is the complete conditional distribution  is even more harder to be extracted.

\end{description}

\section{Interpretability levels }
Inspired by the work of Judea Pearl in which the author suggests three hierarchical levels of modeling, we present three different levels of interpretability \cite{pearl2018theoretical}. 

\begin{description}

\item[$\bullet$ Statistical (Associational) interpretability] Main stream machine learning methods operate in a statistical or model-free mode which only depends upon the observations. Using these traditional statistical based machine learning approaches, we can only answer the interpretability partially. The base for statistical machine learning methods is the association rule \(p(y|x)\) which only considers the observation occurred. We call this type of interpretability as statistical interpretability of machine learning methods. The typical question answered at this level is \textbf{\textit{``what is"}} or \textbf{\textit{``How would seeing ``x" change my belief in ``y" "}}.\\ 

\item[$\bullet$ Causal interventional interpretability] There is serious theoretical limit on the interpretation capabilities of statistical machine learning models because of the absence of causal interventions. There is another level of interpretability which we call interventional causal interpretability that can not be obtained merely on the basis of statistical modeling. We need to model the problem in terms of graphical causal methods which can be represented by an intervention rule \(p(y|do(x), z)\) \cite{pearl2003causality}. The typical question answered at level is \textbf{\textit{``what if"}} or \textbf{\textit{``what if I do ``x",  does that affect the probability of ``y" "}}. The natural dynamics of the variable ``\(x\)" is changed and that is why, any question asked regarding this variable can not be answered from observed data. This kind of interventional interpretation is often required in daily life like what If I take aspirin, will my headache be cured? Or what if we ban the cigarettes?  Causal interventions are not probabilistic, they are beyond probabilities, related to the change in the natural dynamics of the problem. This is very different from \(p(y|x)\), because in the latter, \(x\) assumes some natural distribution in the purely observation setting.

\item[$\bullet$ Counterfactual interpretability] At the top most layer, we have counterfactual, where after observing an outcome, we ask the question \textbf{\textit{``is it because we did some action?"}}. Counterfactuals are difficult to design because of a model a new kind of conditional probability, namely \(p(y_x|x^`,y^`)\). This equation dictates that in a natural setting, we observe \(x^`\) and \(y^`\) which produces certain probability, namely \(y\). Then how would the probability of \(y\) changes to \(y_x\) if we would set \(x^`\) to \(x\). The activities involved in counterfactual probability modeling are imagining and retrospecting. These type of questions can be answered only by using function or structural equation models \cite{pearl2003causality}.
\end{description}

\section{Interpretability as science rather than a tool}

Interpretability is not a monolithic concept but rather defined differently for
various problems and on different levels. Almost all the research involved is mainly focusing on the techniques rather than the notion of what does the term itself mean?  Miller defines it as interpretability is the degree to which human can understand the cause of a machine learning decision\cite{miller2017explanation}. In order to understand the cause of a decision made by machine learning model, one need to have domain knowledge in that specific field. Interpretability without considering a specific task makes no sense. Even for humans, the concept of interpretability is problem specific
and varies from one person to another person for the same problem. As we know that sciences are
preliminary defined by the set of questions rather than by tools. For instance,
astrophysics is discipline that learns the composition of stars, not as the discipline
that use the spectroscopes. We focus on the set of questions and try to answer them using different instruments. New sophisticated tools are developed but with an aim of answering the questions in better way.  Similarly, there is a need to define set of
generic questions that covers different aspects of interpretability. The questions
should be of association as well of of causal nature. They should be flexible enough to be modified for different tasks within the same domain. Then for certain problem, different machine learning interpretability tools should be applied to answer the predefined questions. This will pave a path to discover new set of interpretability methods that will help us understand the specific domain more deeply. By this notion, we will be using machine learning and artificial intelligence models to help us understand about the real world problems rather than relying on the models. Question based approach for interpretability will help us acquiring more knowledge of specific domain, and thus using AI for elevating human understanding of different real world problems rather than eliminating the human intelligence component all together.

\section{Model questions}
The type of questions depends upon specific domain, but still we can devise set of model questions at three levels of interpretability. We formulated few general questions as well as identified them if they belong to causal, association and counterfactual level. These questions along with the identified level are given below. It should be noted that these questions are formulated in the context of classification.\\
\begin{itemize}
\item Which feature/s are the most important ones in the context of classification or prediction? (Association)
\item What is the range of values for selected important features to discriminate a specific class? (Association)
\item How the model came across certain decision? (Statistical Transparency)
\item How sensitive is the class output to a specific feature? (Associational)
\item How does the final class output change if we force specific feature to get some value which is not observed in the data? (Causal Intervention)
\item How effective is a specific feature in resulting a specific class? (Causal Effectiveness)
\item How the final class output would have changed if some feature (or their specific values) had not occurred? (Counterfactual) 
\end{itemize}

\section{Hypothetical framework for interpretability}

As mentioned previously, state of the art machine learning methods like deep learning are producing highly accurate results but suffer from black-box nature. Similarly, decision tee methods are claimed as interpretable with relatively low classification or prediction accuracy. An Ideal framework for high accuracy and interpretability should have the following attributes. 

\begin{description}

  \item[$\bullet$ Performance] The accuracy should be comparable to the state of the art methods.
  \item[$\bullet$ Specifications of interpretability] Questions related to interpretability defined for the problem prior to the training should be answered after training the model.
  \item[$\bullet$ Transparency] The model should be made as simple as possible so that it can be transparent. 
 
\end{description}

\begin{figure}[!h]
\caption{Hypothetical framework for three level interpretability}
\centering
\includegraphics[width=10cm,height=10cm,keepaspectratio]{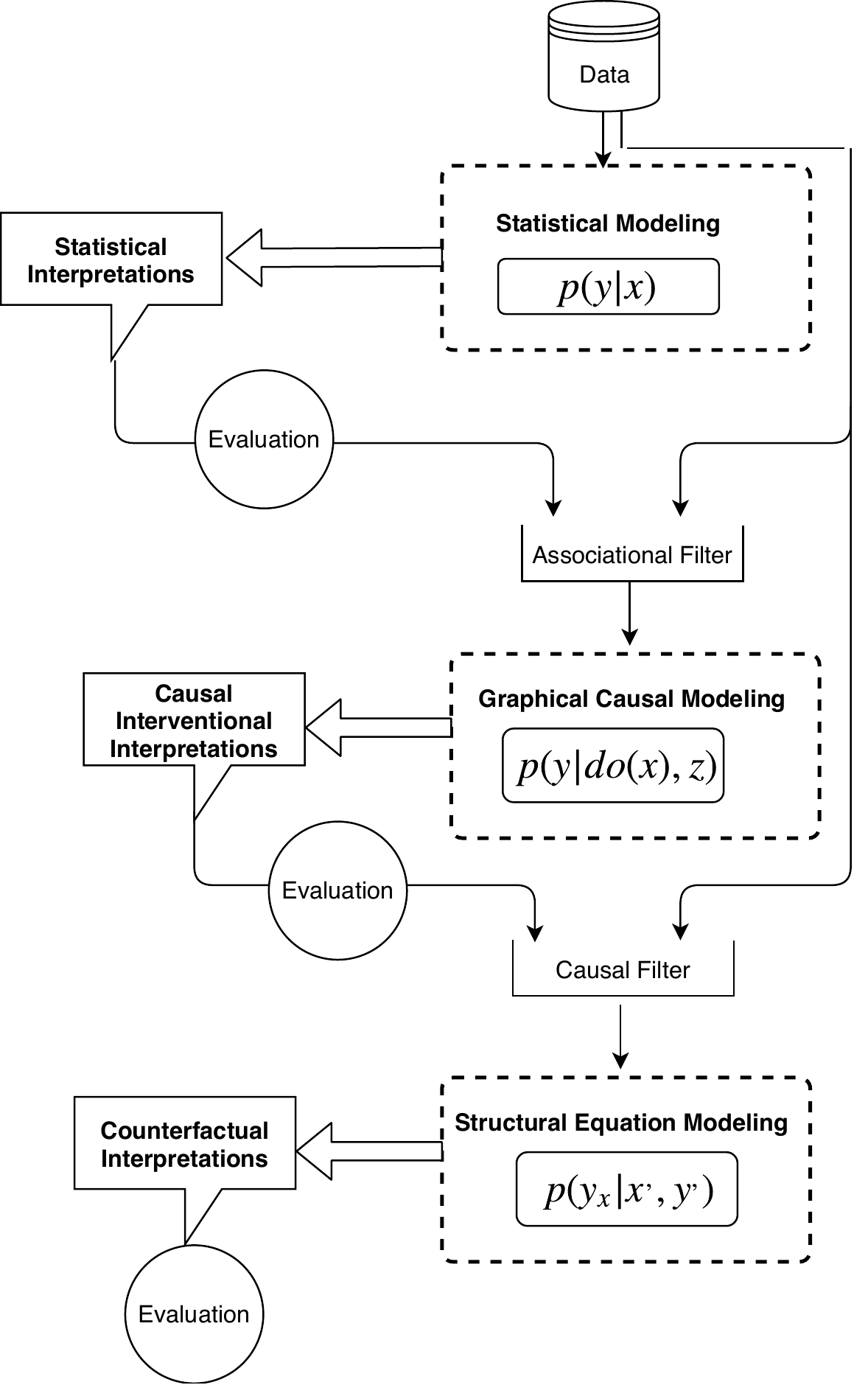}
\label{fig:Framework}
\end{figure}

We present a hypothetical approach to address the questions related to different levels of interpretability as shown in Figure~\ref{fig:Framework}. It should be noted that the performance and transparency are related to the statistical or association based modeling while the interpretability is defined over three levels. The association level model act as a coarse filter to pour the relevant knowledge into interventional level which in turn pours its knowledge to counterfactual level. The main idea is to use three different types of modeling techniques in a way that one compliments the other in an order.

\subsection{Case study}
Let's assume we are dealing with a binary classification problem having full features set such as \(x_{full}=\) $[\![ x_1,x_1,.....x_n, ]\!]$. A deep neural network or random forest classifier is chosen as a statistical model to learn the correlation between the input and output variables. After choosing the statistical model for training and achieving the desired performance, any type of model agnostic method is used to answer the predefined statistical interpretability questions. For instance, the questions like ``which features are the most important ones in the context of classification?", ``What is the range of values for selected important features to discriminate a specific class?", ``How sensitive is the class output to a specific feature?" can be answered at this level. Depending upon the specific problem, other types of association related questions can also be formulated and answered. The answers should be evaluated using the expert knowledge. Now we select the most important features set, say \(x_{corr}\) such that  $x_{corr}\subset\ x_{full}$, which are strongly correlated with the output. 
Though the \(x_{corr}\) features set has proven to be strongly correlated with the output, but in order to find if they are causally associated with the output or not, we can go to the next level of graphical causal modeling. We can do bayesian causal modeling using only the \(x_{corr}\) features set complimented with expert knowledge to answer the causal interventional questions like``How does the final class output change if any specific feature value is to be changed in a way which is not observed in the data?" The statistical interpretations can be used as a source of initial knowledge via associational filter to guide the causal modeling. Now at this level, we have obtained feature set, say \(x_{cause}\) which are not only correlated with the output but also proved to be causation of the output. It should be noted that  $x_{cause}\subset\ x_{corr}\subset\ x_{full}$.
\(x_{cause}\) is then used to do structural equation modeling and the counterfactual questions are answered at this level.

\section{Discussion}

In order to unveil the science of interpretability in machine learning, we have to look at how humans interpret their decisions. If we present a chest x-ray to cardiologist, he might not be able to interpret it. But the same x-ray report can be interpreted by a radiologist. Similarly, interpretability is machine learning should also be considered domain specific subject. Without questions based notion of interpretability, machine learning is not helping the scientists in enhancing their understanding about specific domain. It is also very important to understand the limits of statistical modeling and compliment it with other approaches like causal graphical and structural modeling. The challenge though is to compliment interpretations from one level to another level and formulation of the problem at different levels. By carefully defining the questions we need to ask beyond accuracy and applying different approaches to extract the answers of those questions, we can discover new knowledge. This notion of interpretability is much needed in the high risk and high critical fields with a potential of knowledge discovery like drug design, bioinformatics etc. Problem formulation and evaluation of interpretations is still going to pose many challenges.

\section{Conclusion}

Although there is no absolute well agreed upon
definition of interpretability in literature regarding machine learning methods,
we can still formulate some general rules to interpret the decisions. We
propose a hypothetical hybrid approach which caters the three basic levels of interpretability, i.e., statistical,
interventional causal and counterfactual interpretability.
Statistical interpretability is related to the questions that can be answered
from the already observed data. The inherited association rules enable us to
provide an explanation of the questions of the type ``What is the probability
of certain output if we have observed some specific value of certain variable?"
The second level which is interventional causal interpretability
that can not be dealt with using only the association based
statistical modeling. Causal modeling approaches based on intervention rules
are central to answer questions related to this level of interpretability. The
typical question related to causal interpretability includes ``How will the intervention
of some new unobserved feature affect the final outcome
?". Such intervention
can not be inferred from the observed data, rather it needs causal modeling. The third level is counterfactual which deals with the retrospection and imagining. The association level model act as a coarse filter to pour the relevant knowledge into interventional level which in turn pours its knowledge to counterfactual level. The main idea is to use three different types of modeling techniques in a way that one compliments the other in an order.
Looking at the problem of interpretability in the context of domain specific questions at different levels, we can achieve the end user goal of machine learning methods. This notion of interpretability might be a step towards science based machine learning rather than merely engineering.

\bibliographystyle{named}
\bibliography{ijcai18}

\end{document}